\documentclass[final,5p,times,twocolumn]{elsarticle}

\usepackage{algorithm}
\usepackage{algorithmic}
\usepackage{amsmath}
\usepackage{times}
\usepackage{epsfig}
\usepackage{amsfonts}
\usepackage{amssymb}
\usepackage[table]{xcolor}
\usepackage{multirow}
\usepackage{svg}
\usepackage{stmaryrd}
\usepackage{subcaption}
\usepackage{ulem}
\newcommand{\addSylv}[1]{\textcolor{black}{#1}}
\newcommand{\addSido}[1]{\textcolor{black}{#1}}
\newcommand{\comSylv}[1]{\textcolor{orange}{\textit{#1}}}

\journal{EAAI}

\begin{document}

\begin{frontmatter}



\title{Anomaly-Aware YOLO: A Frugal yet Robust Approach to Infrared Small Target Detection}



\author[1]{Alina Ciocarlan\corref{cor1}} 



\ead{alina.ciocarlan@polytechnique.edu}




\author[2]{Sylvie Le Hégarat-Mascle}

\author[3]{Sidonie Lefebvre}

\affiliation[1]{organization={French Ministerial Agency for Defense AI (AMIAD)},
    city={91120 Palaiseau},
    country={France}}
\affiliation[2]{organization={SATIE, Paris-Saclay University},
    city={91405 Orsay},
    country={France}}

\affiliation[3]{organization={DOTA \& LMA2S, ONERA, Paris-Saclay University},
    city={ F-91123 Palaiseau},
    country={France}}

\cortext[cor1]{Corresponding author}
\begin{abstract}
Infrared Small Target Detection (IRSTD) is a challenging task in defense applications, where complex backgrounds and tiny target sizes often \addSylv{result in} numerous false alarms using conventional object detectors. To overcome this limitation, we \addSylv{propose} Anomaly-Aware YOLO (AA-YOLO), which \addSylv{integrates} a statistical anomaly detection test \addSylv{into} its detection head. By treating small targets as \textit{unexpected} patterns against the background, AA-YOLO effectively controls the false alarm rate. Our approach not only achieves competitive performance on several IRSTD benchmarks, but also demonstrates remarkable robustness \addSylv{in scenarios with} limited training data, noise, and \addSylv{domain shifts}. Furthermore, since only the detection head is modified, our design is highly generic and has been successfully applied \addSylv{across} various YOLO backbones, including lightweight \addSylv{models}. It also provides promising results when integrated into an instance segmentation YOLO. This versatility makes AA-YOLO an attractive solution for \addSylv{real-world} deployments where resources are \addSylv{constrained}. The code will be \addSylv{publicly released}.
\end{abstract}




\end{frontmatter}



\section{Introduction}
\label{sec:intro}

\begin{figure}
    \centering
    \includegraphics[width=1\linewidth]{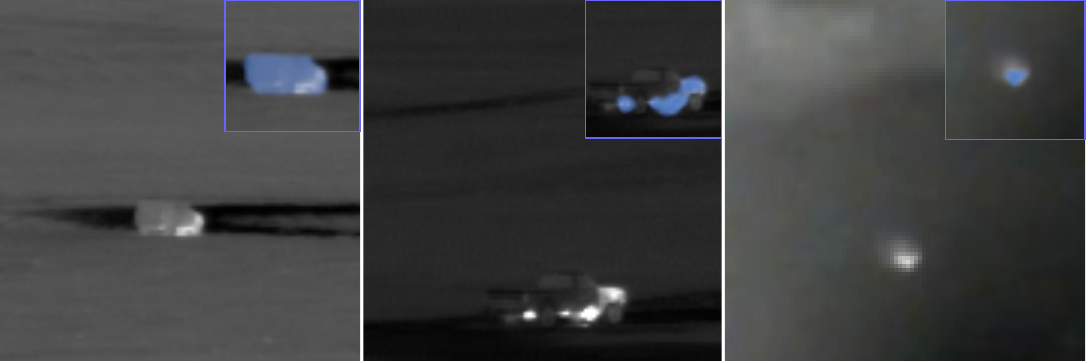}
    \caption{Illustration of the annotation subjectivity in IRSTD. The top-righ corners show in blue the annotation masks provided by~\cite{dai2021asymmetric} for the SIRST dataset.}
    \label{fig:annot_subj}
\end{figure}

Infra\addSylv{R}ed Small Target Detection (IRSTD) is a highly challenging yet critical task in defense, characterized by tiny target sizes, complex backgrounds, and difficult learning conditions. 
Deep learning-based IRSTD methods have been proposed to \addSylv{tackle these} challenges, achieving State-Of-The-Art (SOTA) \addSylv{performance}. These approaches leverage techniques such as dense nested architectures~\cite{li2022dense} or attention mechanisms~\cite{yuan2024sctransnet,hu2025datransnet} to mitigate information loss on small targets and reduce confusion with background elements.
However, current SOTA IRSTD methods \addSylv{face} limitations due to their reliance on segmentation networks. \addSylv{One} major issue is that their evaluation is heavily influenced by subjective annotations, as \addSylv{illustrated} in Figure~\ref{fig:annot_subj}. Specifically, annotators may label entire vehicles or \addSylv{highlight} the most salient \addSylv{regions} with high \addSylv{InfraRed (}IR\addSylv{)} responses, leading to contradictory training signals. These inconsistencies can significantly \addSylv{affect both} the training process and the pixel-level \addSylv{evaluation} metrics. Furthermore, \addSylv{segmentation-based} methods often suffer from i)~object fragmentation when binarizing feature maps, and ii)~adjacency issues, where two \addSylv{nearby targets} are \addSylv{mistakenly} detected as \addSylv{a single object}. These issues affect counting accuracy, especially in critical domains like civil security.

Object detection algorithms \addSylv{such as} YOLO~\cite{redmon2016you} \addSylv{help} mitigate this risk by explicitly localizing objects through bounding box regression, with faster inference times. Although annotation subjectivity has \addSylv{less} impact when switching from pixel-level to object-level annotations, \addSylv{it remains non-}negligible for small objects. Indeed, \cite{cheng2023towards} highlight that \addSylv{even} minor localization errors can severely impact Intersection over Union (IoU) metrics for small objects, impairing \addSylv{both} training and evaluation for YOLO networks. 
Recent approaches~\cite{yang2024eflnet,yang2025pinwheel} address these issues by reducing the \addSylv{influence} of IoU loss on small objects and \addSylv{proposing} alternative loss functions. While effective, SOTA methods often result in complex, task-specific \addSylv{models} that may not adapt well to real-world scenarios with limited resources.

\addSylv{In this paper, we advocate for a complementary direction: \textbf{detecting small IR targets as statistical anomalies with respect to the background}.} 
By treating small targets as \textit{unexpected} patterns against the background, we \addSylv{adopt a statistical} hypothesis testing framework \addSylv{where the background distribution is modeled as the null hypothesis}. 
\addSylv{Importantly, this test is performed in latent space and requires minimal assumptions on background structure. We rely on a pragmatic exponential modeling, grounded in the maximum entropy principle~\cite{jaynes1957information}, to derive an interpretable objectness score that tightly controls false alarms.}
Unlike traditional object detectors, our approach explicitly models the \textit{unexpectedness} of small objects in the latent space, \addSylv{enabling both~i) effective anomaly identification and~ii) explicit control of} the false alarms. 
Our method achieves \addSylv{strong} performance under different types of frugality: 

\begin{itemize}
    \item \textit{Data frugality}: our method \addSylv{retains} at least $90\%$ of the \addSylv{full-}performance SIRST dataset~\cite{dai2021asymmetric}\addSylv{, even when trained with} only $10\%$ of the data. 
    \item \textit{Computational frugality}: \addSylv{even when} using lighter networks, our approach is competitive with SOTA, making it suitable for deployment on resource-constrained devices.
    \item \textit{Generic design}: our method is \addSylv{easy} to implement, requiring only \addSylv{a simple} modification of the detection head. This makes our approach highly adaptable and user-friendly.
\end{itemize}

For these reasons, our Anomaly-Aware YOLO (AA-YOLO) \addSylv{is a compelling} solution for real-world applications where \addSylv{computational and data} resources are \addSylv{limited}. Our \addSylv{main} contributions can be summarized as follows:
\begin{enumerate}
    \item We propose a simple yet effective YOLO detection head called the Anomaly-Aware Detection Head (AADH) that integrates statistical anomaly testing to improve IRSTD. Our method provides an anomaly-informed objectness score that empirically suppresses background.
    \item By adding our AADH module to several conventional YOLO backbones, especially lightweight ones, \addSylv{we close} the performance gap with SOTA \addSylv{methods}. Notably, AA-YOLOv7t \addSylv{achieves} SOTA results on famous IRSTD benchmarks while having six times fewer training parameters \addSylv{than} EFLNet. It also gives promising results when integrated into an instance segmentation YOLO.
    \item AADH significantly improves robustness in frugal settings, transfer learning, inference on noisy data, and operational contexts.
    \item Lastly, our AADH facilitates the choice of a detection threshold by constraining all background values to zero. 
\end{enumerate}

\section{Related works}
\subsection{IRSTD methods}
\label{sec:related_IRSTD}
The specificity of IRSTD lies in the extremely small size and low contrast ratio \addSylv{of the targets}, making them \addSylv{particularly} difficult to detect against complex backgrounds. Researchers have developed various deep learning-based strategies that surpass traditional methods for IRSTD. These methods fall into two main categories.

\paragraph{Segmentation methods}

Segmentation networks are particularly popular among IRSTD methods due to their inherent ability to extract \addSylv{fine-grained} features. To further \addSylv{enhance} small target detection \addSylv{and reduce confusion}  with background clutter, \cite{zhang2023attention, yuan2024sctransnet, hu2025datransnet} introduce attention mechanisms in their backbones. Specifically, RDIAN~\cite{sun2023receptive} integrates convolution-based multidirection\addSylv{al} guided attention mechanisms to enhance small target features in deep layers, while SCTransNet and DATransNet rely on the recent Transformer backbones to improve global \addSylv{scene understanding}. An important issue with deep neural networks is their pooling layers, which degrade the amount of information for small targets. To mitigate this,~\cite{li2022dense} introduce DNANet, a dense-nested U-shaped backbone that leads to great performance on widely used IRSTD benchmarks. Other approaches \addSylv{attempt to resolve the limitations} of IoU-based metrics for small objects by introducing alternative loss functions. For instance, MSHNet~\cite{liu2024infrared} propose\addSylv{s} a location\addSylv{-} and scale\addSylv{-}sensitive loss, which significantly improves performance when combined with a simple U-Net architecture.

However, these methods are not robust to fragmentation issues that arise when binarizing segmentation maps, which \addSylv{can distort object counts}. Additionally, they are prone to adjacency errors, where two adjacent objects are incorrectly merged. This leads to the exploration of object detectors with bounding box regression for IRSTD.

\paragraph{Object detection methods}
Object detection involves detecting objects of interest within an image and identifying their locations with bounding boxes. Several deep learning approaches have been proposed for this task, including the popular YOLO framework~\cite{redmon2016you}. While YOLO detectors achieve great performance in various applications with low \addSylv{inference} time, they struggle to detect small objects due to two key factors: (1)~class imbalance between target and background samples; and (2)~low tolerance to bounding box localization errors for small objects, \addSylv{as even minor offsets can cause large drops in IoU}. To address these issues, dedicated loss functions~\cite{yang2024eflnet} have been introduced to increase the \addSylv{relative importance} of minority samples. Additionally, \cite{yang2025pinwheel} propose to reduce the weight of the IoU metric in the loss for small objects through their Scale Dynamic (SD) loss. The authors of EFLNet~\cite{yang2024eflnet} go even further by replacing IoU-based loss with the Normalized Wasserstein Distance, as proposed in~\cite{xu2022rfla}. EFLNet's performance on IRSTD benchmarks is particularly noteworthy, \addSylv{surpassing} SOTA segmentation-based methods and setting \addSylv{a new performance} standard for YOLO-based networks for IRSTD.

It is worth noting that existing papers often only partially address the various aspects of frugality. While some lightweight detectors for IRSTD have been proposed~\cite{ma2023dmef,nguyen2025lw}, their robustness in frugal settings is rarely evaluated. Our paper aims to bridge these gaps by introducing a simple\addSylv{,} resource-efficient\addSylv{,} yet versatile approach that not only achieves SOTA performance for IRSTD but also demonstrates robustness.

\subsection{Anomaly detection}

The small size, lack of structure and scarcity of the targets make them natural candidates for anomaly detection. Indeed, \addSylv{this task aims to identify} rare events that deviate from the standard and for which \addSylv{only} limited samples \addSylv{are available}. According to~\cite{ruff2021unifying}, these methods can \addSylv{fall} into four \addSylv{main categories}: distance-based methods (e.g., isolation forest or k-nearest neighbors), reconstruction-based methods, one-class classification methods, and probabilistic methods. 
Reconstruction-based methods detect anomalies \addSylv{as inputs} that are challenging to reconstruct accurately~\cite{baldi2012autoencoders, nguyen2024variational}. 
One-class classification methods, such as Deep SVDD~\cite{ruff_deep_2018}, \addSylv{train solely on normal samples and learn a compact representation or boundary around them, flagging outliers as anomalies}.
Probabilistic methods \addSylv{instead aim to model} the probability density function of the normal data\addSylv{, identifying anomalies as samples with low likelihood}. \addSylv{A notable example is t}he Reed-Xiaoli algorithm \addSylv{widely} used in hyperspectral anomaly detection~\cite{kwon2005kernel}. 
Note that hypothesis testing also falls \addSylv{into} this category. 
Despite its \addSylv{potential}, \addSylv{the integration of} anomaly detection into end-to-end supervised learning frameworks for IRSTD remains limited. For example,~\cite{shi2020infrared} treat small targets as ``noise'' in an infrared background, \addSylv{reframing} the detection problem to a denoising task. Similarly, \cite{deng2024bemst} model the background \addSylv{in an unsupervised manner} using VAEs and then applies a supervised detection algorithm to the resulting difference image. While promising, we \addSylv{argue} that explicitly integrating the notion of \textit{anomaly} within a supervised detection pipeline to guide feature extraction could improve robustness and accuracy, particularly in challenging scenarios where traditional methods often fail.

\section{Proposed method}
Our objective is to improve small target detection, especially under challenging conditions, by introducing an anomaly-based prior in\addSylv{to} a YOLO-type network. To this \addSylv{end}, we train \addSylv{the} network not to estimate the decision boundaries between target and non-target data points, as is conventionally done in detection networks, but \addSylv{rather to identify deviations} from the background model. This is done by constraining the feature extraction \addSylv{using} a probabilistic anomaly detection method. Integrating this statistical criterion \addSylv{into} the training loop of a YOLO network ensures that, after convergence, only the targets \addSylv{violate the learned} background model.

\subsection{Formulation of the statistical anomaly testing}
\label{sec:anomaly_testing}
Our contribution consists in re-estimating the objectness scores predicted by the YOLO detection head for each bounding box \addSylv{using} multiple statistical tests rejecting the null hypothesis $H_0$. To this \addSylv{end}, we consider the $N$ voxels from the final feature map of dimension $H\times W \times C$, where $H$, $W$ and $C$ are the height, width and number of channels of the feature map, respectively. Each voxel $v_k \in \mathbb{R}^{1\times 1\times C }$ is represented by a $C$-dimensional random variable $X_k=(X_{k,1}, ..., X_{k,C})$, where $X_{k,1}, ..., X_{k,C}$ are assumed \addSylv{to be} independent and identically distributed (i.i.d.). 
\addSylv{Note that the i.i.d. assumption across feature channels is a simplification and does not strictly reflect the dependencies learned by CNNs. However, in our context, this assumption does not aim to describe the exact generative process of feature activations, but to provide a tractable and interpretable null hypothesis for anomaly detection.}
According to the null hypothesis $H_0$ (to be rejected), a given voxel belongs to the background class. Our focus here is on controlling the Type I error rate, i.e. the probability of rejecting $H_0$ \addSylv{when} it is \addSylv{actually} true. \addSylv{When performing} multiple hypothesis tests, \addSylv{this} can be framed as controlling the Family-Wise Error Rate (FWER). Let $\theta \in \mathbb{R}^{*}$ be the rejection threshold, $F$ the testing function, and $\mu$ a measure. The FWER \addSylv{at} confidence \addSylv{level} $\alpha\in \mathbb{R}^*$ is defined as:

\begin{equation}
    \mathbb{P}_{H_0} (\exists k \in \llbracket 1,N\rrbracket,  F (\mu(X_k)) < \theta) < \alpha.
\label{eq:type_1_error}
\end{equation}

To compute the FWER, we define $F$ as the p-value function. Given an observation $x_k$, the p-value is expressed as:

\begin{equation}
    F(\mu(x_k)) = \mathbb{P}_{H_0}(\mu(X_k) \geq \mu(x_k)).
    \label{eq:NFA_form}
\end{equation}

Let us now focus on our problem to explicitly define $H_0$ and $F$\addSylv{. First, note that our objective is not to perfectly fit the latent distribution of background voxels, but to select a plausible null hypothesis enabling effective anomaly discrimination, which is sufficient for our operational goals}. \addSylv{M}odeling IR background distributions can be challenging due to dynamic textures\addSylv{. However, o}ur anomaly test operates in latent space, where background voxel features \addSylv{tend to} cluster near zero after the ReLU activation, regardless of input image complexity.
In this latent space, assuming an exponential distribution is well \addSylv{justified} since it is the maximum entropy distribution \addSylv{among} non-negative variables with fixed mean. \addSylv{Thus, according to the principle of maximum entropy~\cite{jaynes1957information}, this makes it the minimally biased choice, introducing only the known constraints (here, non-negativity and empirical mean). 
Note that this assumption is} \addSylv{
pragmatically validated, as alternative hypotheses (e.g., Gaussian) lead to inferior detection performance (see our ablation study in Section~\ref{sec:ablation})}\addSylv{, suggesting that the exponential assumption better aligns with the structure of the latent features.}
We therefore \addSylv{define} the null hypothesis $H_0$ \addSylv{as assuming that each voxel} follows a $C$-dimensional exponential distribution $\mathcal{E}$ with parameter $\boldsymbol{\Lambda} = [\lambda_1, ..., \lambda_C]^\mathbf{T}$. 

Then, two aggregated measures $\mu$ admitting closed-form distributions under the exponential hypothesis can be considered:
\begin{enumerate}
    \item $\mu_1(X_k) = \min \{X_{k,1}, ..., X_{k,C}\}$ -- In this case, the minimum of the channel values follows an exponential distribution \addSylv{with} parameter $\lambda_{\mu_1}=\sum_{i=1}^C\lambda_i$. The test function $F$ then simplifies to:  
    \begin{equation}
        F(\mu_1(x_k))= e^{- \lambda_{\mu_1}\addSylv{\cdot} \min \{x_{k,1}, \dots, x_{k,C}\}}.
    \end{equation}
In practice, each $\lambda_i$ is estimated as the reciprocal of the average \addSylv{activation} value across the spatial dimension of the $i^{th}$ channel in the feature map. 
    \item $\mu_2(X_k) = \sum_{i=1}^C X_{\addSylv{k,}i} $ -- Assuming that all $\lambda_i$ are equal to $\lambda_{\mu_2}$, the sum follows an Erlang distribution with shape parameter equal to the number of summed variables ($C$) and rate parameter equal to that of the common exponential distribution of parameters $\lambda_{\mu_2}$, leading to: 
    \begin{equation}
    F(\mu_2(x_k)) = \frac{\Gamma(C, \lambda_{\mu_2} \sum_{i=1}^C x_{k,i})}{\Gamma(C)},
    \label{eq:NFA_form_gamma}
\end{equation}
 where $\Gamma(\cdot)$ and $\Gamma(\cdot,\cdot)$ are the Gamma and upper incomplete Gamma functions, respectively. Here, $\lambda_{\mu_2}$ is computed as the reciprocal of the average \addSylv{activation} value \addSylv{across} all \addSylv{voxels in} the feature map.
\end{enumerate}

\begin{figure}
    \centering
    \includegraphics[width=1\linewidth]{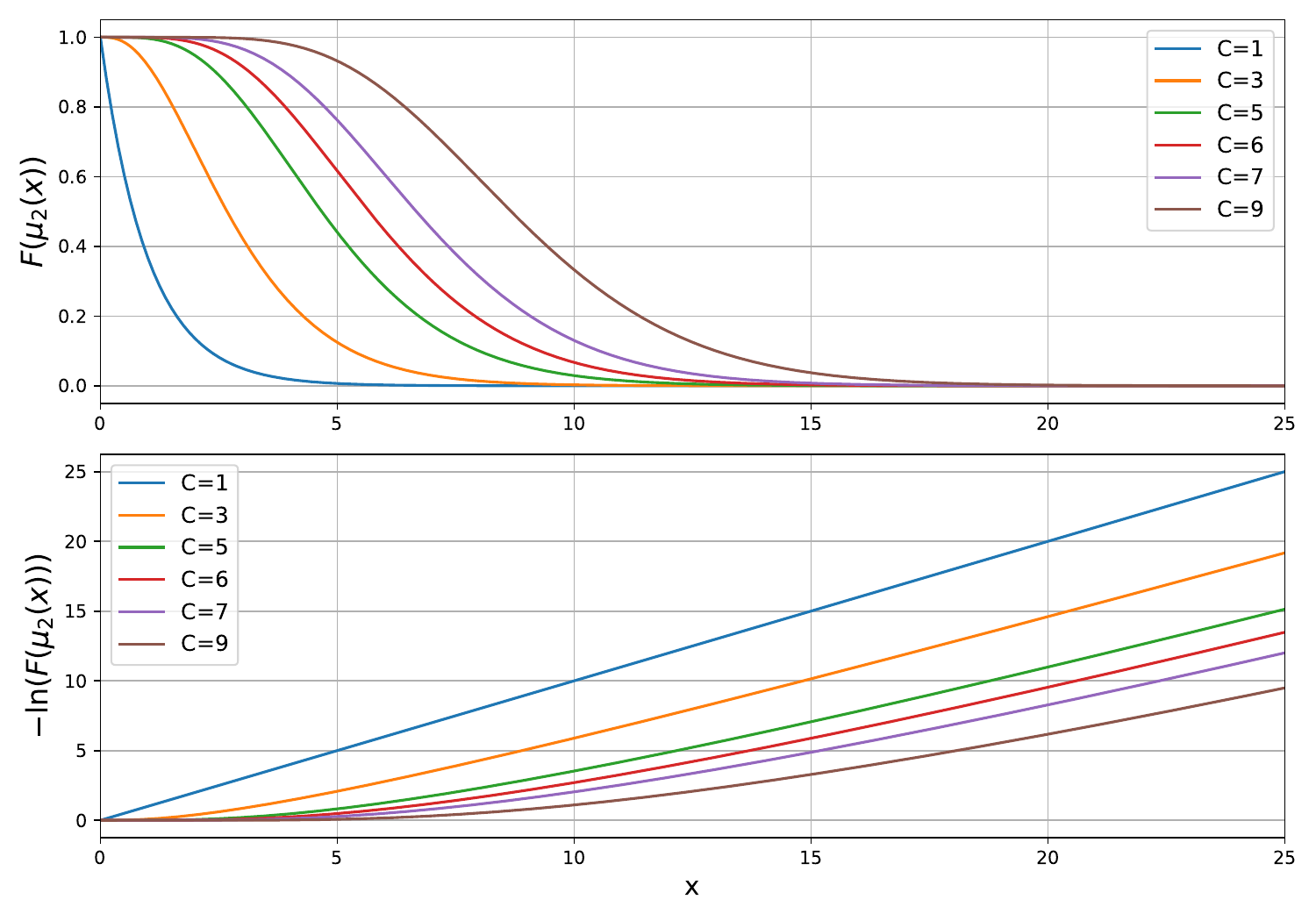}
    \caption{Variations of $F_{\mu_2}$ and $-\ln{F_{\mu_2}}$.}
    \label{fig:F_variations}
\end{figure}
Our ablation study, detailed in Section~\ref{sec:ablation}, shows that the measure $\mu_2$ (sensitive) outperforms $\mu_1$ (conservative). This \addSylv{superiority may stem from a key} limitation of $\mu_1$, which \addSylv{implicitly} assumes that all channels \addSylv{must simultaneously exhibit target-like behavior}, i.e. no channel $i$ exhibits a background-like $x_{k,i}$ value, once one channel does. 
Such an assumption conflicts with the goal of extracting \addSylv{diverse and complementary} features \addSylv{across the channels, potentially degrading the overall discriminative power of the representation}. Consequently, we \addSylv{adopt} $\mu_2$ \addSylv{for the rest of the study}. 
\addSylv{The shape} of our test function \addSylv{is} displayed in Figure~\ref{fig:F_variations}\addSylv{. We report scores} in terms  of \textit{significance}~\cite{HegaratMascle2019}, i.e. $-\ln{F(\mu(x_k))}$, to \addSylv{enhance} interpretability.
To our knowledge, \addSylv{the use of} an exponential hypothesis \addSylv{in this detection setting} has not been previously explored. Its \addSylv{key} advantage lies in its ability to explicitly \addSylv{push} background \addSylv{activations towards} zero, \addSylv{thereby} simplifying the selection of detection thresholds under \addSylv{real-world} conditions.

\subsection{Integration within the YOLO framework}
\paragraph{Architecture overview}
\begin{figure*}
    \centering
    \includegraphics[width=0.99\linewidth]{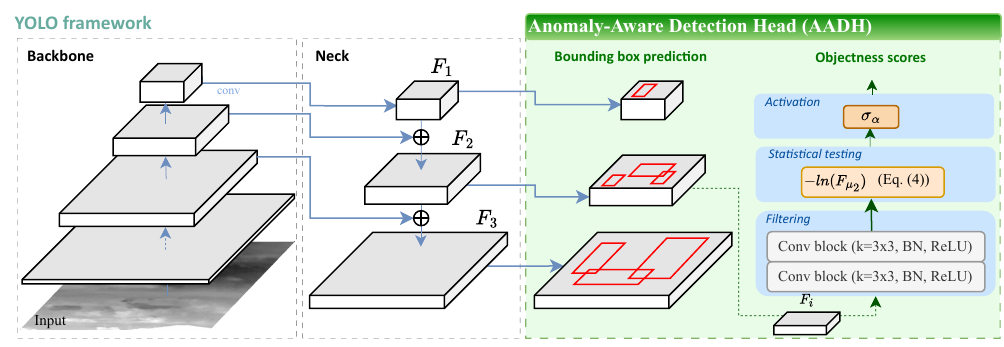}
    \caption{AA-YOLO architecture. The input first passes through a generic YOLO backbone to extract high-level features. Then, the conventional YOLO detection head is replaced by our Anomaly-Aware Detection Head (AADH), which enhances small object features by integrating statistical anomaly testing to estimate objectness scores. For simplicity, we illustrate the application of our AADH on $F_2$-level feature maps only, although in practice AADH is applied to all detection scales: $F_1$, $F_2$ and $F_3$.}
    \label{fig:archi}
\end{figure*}
The overall architecture is illustrated in Figure~\ref{fig:archi}. The input image is first processed by a \addSylv{standard} YOLO network, such as YOLOv7 or YOLOv9. To \addSylv{ensure} the generality of our approach, we only modify the detection head of the YOLO network. Specifically, we \addSylv{decouple} the prediction of the objectness scores from that of the bounding box coordinates and class scores.
We then re-estimate the objectness score using our Anomaly-Aware Detection Head (AADH), which integrates the \addSylv{proposed statistical anomaly test}. This process yields new objectness scores that range from $0$ to $1$, \addSylv{now reflecting the notion of anomaly with respect to the learned background distribution}. 
The entire network, including the AADH, is trained end-to-end using the Mean Square\addSylv{d} Error (MSE) loss for the objectness scores. \addSylv{We refer to the resulting architecture} as \textbf{AA-YOLO} (Anomaly-Aware YOLO). \addSylv{When s}pecifying the used backbone, e.g., YOLOv7, the network is called AA-YOLOv7.

\paragraph{Anomaly-Aware Detection Head}

As illustrated in Figure~\ref{fig:archi}, the AADH consists of three main components. First, a spatial filtering block is designed to capture and gather \addSylv{relevant spatial context around the central voxel of each predicted bounding box}. This \addSylv{block includes} two convolutional layers with a $3\times 3$ kernel size, each followed by a Batch Normalization layer and  ReLU activation. The output is a \addSylv{$C$-channel} feature map, which then undergoes the statistical test $-\ln(F_{\mu_2})$ (where $F_{\mu_2}$ is given by Eq.~\eqref{eq:NFA_form_gamma}).
\addSylv{The behavior of the function $F_{\mu_2}$ (cf. Figure~\ref{fig:F_variations})} highlights that the choice of $C$ \addSylv{influences} the sharpness of $F$. Our ablation study suggests setting $C=8$ \addSylv{for optimal detection} performance. 
To ensure that the \addSylv{output} scores \addSylv{from} the statistical test fall between $0$ and $1$, we employ a scaled and zero-centered sigmoid activation function, parameterized by $\alpha$, which effectively handles the asymmetry of the output scores. 
\addSylv{This} activation function \addSylv{is defined} as $\sigma_{\alpha}(x)=\frac{2}{1+e^{-\alpha x}}-1$. Given that our statistical test produces scores that slowly increase (almost linearly, as shown in Figure~\ref{fig:F_variations}) and reach particularly high values for targets, a smaller $\alpha$ (below 1) helps to stretch the sigmoid curve, allowing it to better capture the nuances of our output scores. Empirically, we find in our ablation study that setting $\alpha=0.001$ leads to optimal performance, as it provides a suitable trade-off between sensitivity and specificity.

\section{Experiments and results}

\subsection{Experimental set-up}
\paragraph{Datasets} 

To evaluate our methods, we \addSylv{rely on} two widely-adopted benchmarks for IRSTD: the SIRST dataset~\cite{dai2021asymmetric} and the IRSTD-1k dataset~\cite{zhang2022isnet}. The SIRST dataset, one of the first publicly released real-image infrared small target datasets, is a commonly used reference in the literature. It consists of $427$ mono-spectral infrared images with a resolution of $256\times256$ pixels. 
To \addSylv{assess} our methods in frugal \addSylv{learning} settings, we train on a subset of $25$ images from the SIRST dataset, which represents less than $10\%$ of the total dataset. To ensure the independence of our results from the training set, we \addSylv{randomly} select three non-overlapping sets of $25$ images and report the average results across these sets.
In addition to SIRST, we also consider the challenging IRSTD-1k dataset, a recently published benchmark that provides a larger collection of $1000$ images with a higher resolution of $512\times512$ pixels. Both datasets are divided into training, validation, and test sets with a ratio of $60:20:20$, and we use the splits provided by~\cite{dai2021asymmetric} for SIRST and by~\cite{yang2024eflnet} for IRSTD-1k. To \addSylv{meet} the input \addSylv{size} requirements of YOLO networks, all images are upscaled to $640\times640$ using bicubic interpolation.

\paragraph{Baselines}
We evaluate our approach using several YOLO-based architectures with varying sizes, including YOLOv7 and \addSylv{lightweight} version\addSylv{s such as}  YOLOv7-tiny (YOLOv7t) or YOLOv9-tiny (YOLOv9t).
We also combine our approach with the loss and module proposed in~\cite{yang2025pinwheel}. \addSylv{We further explore the applicability of our method beyond detection by integrating AADH into a standard} instance segmentation network, namely YOLOv5-seg.
We compare our methods not only with standard YOLO baselines without AADH but also to advanced SOTA methods, including segmentation-based 
\addSido{approaches like DNANet~\cite{li2022dense}, AGPCNet~\cite{zhang2023attention}, SCTransNet~\cite{yuan2024sctransnet}, SIRST-5K~\cite{lu2024sirst}, MSHNet~\cite{liu2024infrared}, DATransNet~\cite{hu2025datransnet}, and YOLO-based detectors like EFLNet~\cite{yang2024eflnet} and YOLO+PConv+SD~\cite{yang2025pinwheel}.}
All YOLO baselines are trained from scratch using Nvidia V100 or A100 GPU for $600$ epochs, with SGD optimizer and a batch size equal to $16$. \addSylv{Additional} training details for all methods \addSylv{are reported} in the~\ref{sec:training_param}. Note that our comparison focuses on SOTA deep learning-based methods for IRSTD, given that traditional approaches have been surpassed~\cite{li2022dense}.

\paragraph{Evaluation metrics}
Our evaluation \addSylv{relies} on standard object-level metrics, namely the F1 score and the Average Precision (AP), i.e. the area under the precision-recall curve. To determine true positives, we \addSylv{adopt} a relaxed criterion: a detected object is considered a true positive if its IoU with the ground truth is at least $5\%$. \addSylv{This} low \addSylv{threshold} is used to avoid penalizing results in cases where annotations are subjective, as presented in the Section~\ref{sec:intro}. We also present the AP$_s$ metric, which computes the AP for very small objects, i.e. those with a surface area of less than $5 \times 5$ pixels.  \addSylv{We additionally report} the IoU for instance segmentation methods, which can be compared to the SOTA segmentation IoU reported in the~\ref{sec:additional_res}. Finally, to gain a deeper understanding of the methods' sensitivity and detection capability, we provide their precision, recall, and false alarm rates per image in the~\ref{sec:additional_res}. Note that for segmentation methods, we convert the predicted segmentation maps to object-level predictions using the code provided by~\cite{li2022dense}, which employs morphological operators.

\subsection{New SOTA results on two IRSTD benchmarks}
\label{sec:results}
\bgroup
\def\arraystretch{1.15}
\begin{table}[] 
\footnotesize
  \centering
\begin{tabular}{p{2.3cm}llllll} 
  \hline
     \multirow{2}{*}{\textbf{Method}}  &  \multicolumn{3}{c}{\textbf{SIRST} } & \multicolumn{3}{c}{\textbf{IRSTD-1k} }\\\cline{2-7}
      
        &  \textbf{F1 }  &$\textbf{AP}$ &$\textbf{AP$_{s}$}$& \textbf{F1 }  & $\textbf{AP}$  &$\textbf{AP$_{s}$}$  \\
     \hline
     
     \hline
     \multicolumn{6}{l}{\textcolor{darkgray}{\textit{SOTA}}}  \\
     ACM &  95.4 & 95.2  &87.1 & 90.9 & 88.2 & 77.8 \\   
  AGPCNet &  93.8 & 92.2 &93.0 & 91.1 & 88.9 & 74.1  \\
   DNANet &  97.1&98.4&96.1 &  90.7 & 87.0 &  79.1\\
     RDIAN & 95.9 & 93.8 &  86.3 & 86.7 & 84.5 & 67.3 \\
     SCTransNet & 95.4 & 95.9 & 92.9 & 91.9 & 90.8 &  87.0 \\
     SIRST-5K & 96.8 & 98.1 & 95.2 & 90.1 &89.3 &  86.7\\
     MSHNet & 94.8 & 95.6 & 92.3 & 92.0 & 91.1 & 88.8 \\
    DATransNet &  93.5& 92.3& 74.9 & 89.1 & 86.7 &  73.2\\
   EFLNet & 96.9 & 98.3  & 97.6 & \underline{92.5} & \underline{96.5} &  \underline{89.8} \\
    YOLOv7t +PConv +SD & 96.8 &97.8 & 97.1 & 87.8 & 93.2 & 86.1 \\   
     \hline
   \multicolumn{6}{l}{\textcolor{darkgray}{\textit{YOLO baselines}}}  \\
YOLOv7  &  96.9 & 98.2 & 97.1 &  \textbf{92.6} & 95.2 & 87.1 \\
YOLOv7t &  96.3& 98.2 & 97.1 & 85.3 & 91.5 & 83.6\\
YOLOv9t & 94.5 & 97.0 &  97.3 & 90.9 & 95.4 &  83.9 \\
     
      \hline
    \multicolumn{6}{l}{\textcolor{darkgray}{\textit{Our methods - Object detection}}}  \\
          
          AA-YOLOv7t +PConv +SD &  96.9 & 98.0 &  97.0 & 91.1 & 96.0 & 89.5\\   
          AA-YOLOv7  & \textbf{97.9} & \underline{98.5} &  \textbf{98.0} & 91.2 & 95.6 & 86.3 \\
           AA-YOLOv7t  & \textbf{97.9} & \underline{98.5}  & \underline{97.9} & \underline{92.5} & \textbf{96.6} & \textbf{90.9} \\
    
           AA-YOLOv9t & \underline{97.4} & \textbf{98.6} &  97.8 & 91.4 & 95.8 &87.9\\
         \hline 
    
           \textcolor{darkgray}{\textit{Instance segm.}} &  \textbf{F1 }  &$\textbf{AP}$ &$\textbf{IoU}$& \textbf{F1 }  & $\textbf{AP}$  &$\textbf{IoU}$  \\\cline{2-7}
     YOLOv5-seg & 95.8 & 98.2 &  56.5 & 87.6 & 91.9 & 34.8 \\
     AA-YOLOv5-seg &  96.9 & \textbf{98.6} & \textbf{72.4} & 90.2 & 88.9 &  \textbf{66.8}\\
             \hline
  \end{tabular}
  \caption{Object-level results obtained on SIRST and IRSTD-1k datasets. Best results are in bold, and second best ones are underlined. }
  \label{tab:sirst_yolor50_NFA}
\end{table}
\egroup
\bgroup
\def\arraystretch{1.15}
\begin{table}[h] 
\footnotesize
  \centering
\begin{tabular}{lcc} 
  \hline
     \multirow{2}{*}{\textbf{Method}}  &  \textbf{SIRST} & \textbf{IRSTD-1k}\\\cline{2-3}
      
        &  \textbf{IoU $\uparrow$ }  &$\textbf{IoU $\uparrow$}$  \\
     \hline
     
     \hline
     \multicolumn{2}{l}{\textcolor{darkgray}{\textit{SOTA}}}  \\
     ACM & 63.5 &60.3 \\   
  AGPCNet & 62.8 &62.1 \\
   DNANet &  65.3 & 62.1\\
     RDIAN &  64.1 &61.3\\
     SCTransNet & \textbf{76.6} & 60.0 \\
     SIRST-5K & 72.1 & 61.9 \\
     MSHNet & 70.7 & 58.4 \\
    DATransNet &  63.5 & \underline{62.4}\\
    \hline
     \multicolumn{2}{l}{\textcolor{darkgray}{\textit{YOLO-based instance segmentation}}}  \\
     YOLOv5-seg & 56.5 & 34.8 \\
     AA-YOLOv5-seg (ours) & \underline{72.4} &  \textbf{66.8}\\
             \hline
  \end{tabular}
  \caption{Pixel-level results obtained on SIRST and IRSTD-1k datasets. The best results are in bold, and the second best results are underlined. \comSylv{}}
  \label{tab:pixel_res}
\end{table}
\egroup

\begin{figure}
    \centering
    \includegraphics[width=\linewidth]{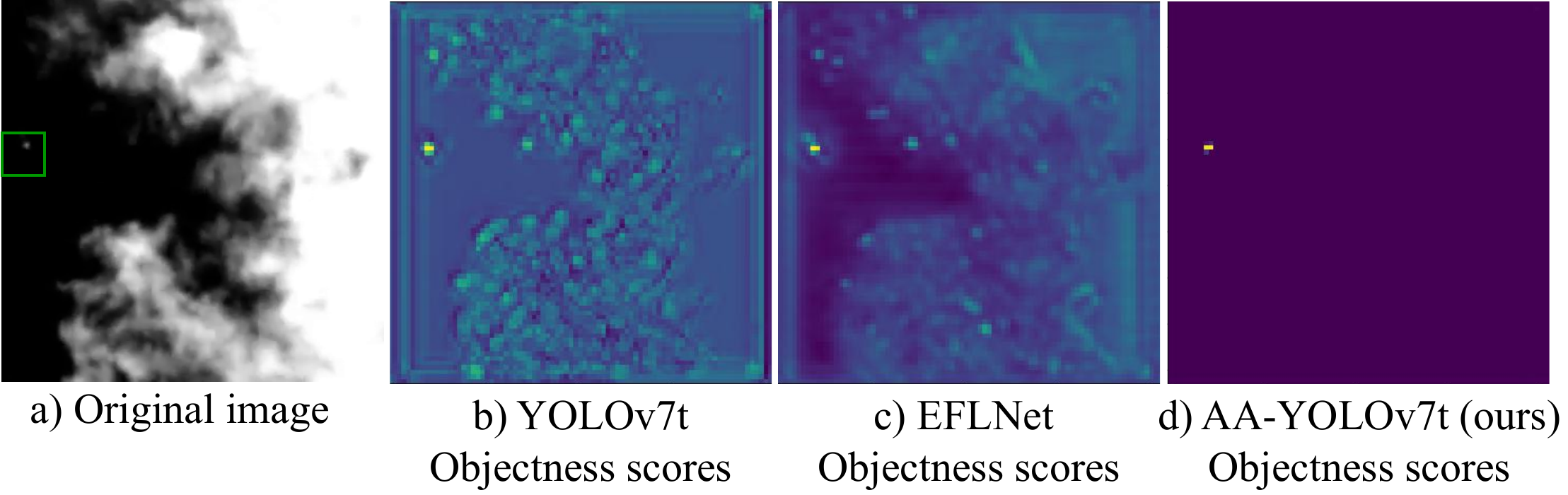}
    \caption{Objectness score maps for different methods.}
    \label{fig:obj_scores_map}
\end{figure}
\begin{figure}
    \centering
    \includegraphics[width=1.02\linewidth]{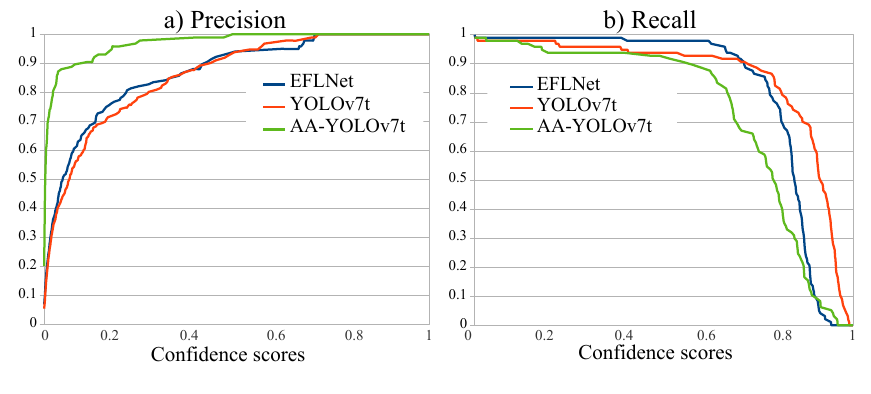}
    \caption{Precision and recall curves obtained on the SIRST dataset.}
    \label{fig:pr_curves}
\end{figure}

\paragraph{Quantitative results}
Table~\ref{tab:sirst_yolor50_NFA} demonstrates that our AADH module \addSylv{consistently improves (with only rare exceptions) and} uniformizes the object-level performance of various YOLO baselines on the SIRST and IRSTD-1k datasets, regardless of the encoder type, number of training parameters (provided in the Table~\ref{tab:comput_res}), or initial performance levels. 
For instance, our smallest backbone YOLOv9t shows a $2.9\%$ increase in F1 on the SIRST dataset and a $4.0\%$ increase in AP$_s$ on the IRSTD-1k dataset when integrating our AADH. 
Compared to SOTA methods, our approach i)~outperforms segmentation-based methods, ii)~benefits the method proposed by~\cite{yang2025pinwheel}, and iii)~is competitive with EFLNet. Notably, our best\addSylv{-performing configuration}, AA-YOLOv7t, improves upon EFLNet by $1.1\%$ in AP$_s$ on IRSTD-1k while having six times fewer training parameters and a frugal design.

Last but not least, the results obtained with AA-YOLOv5-seg are \addSylv{particularly promising}, as it performs competitively with SOTA segmentation methods on the SIRST dataset, achieving similar object and pixel-level metrics. Although it slightly underperforms the best segmentation methods on object-level metrics for IRSTD-1k, it significantly outperforms them in terms of IoU (with a margin of more than $4\%$, as shown in Table~\ref{tab:pixel_res}).
It is worth noting that our AADH significantly enhances the pixel-level performance of YOLOv5-seg. This improvement can be attributed to the challenges posed by the IRSTD task, where YOLOv5-seg struggles to simultaneously learn segmentation and bounding box regression. Adding our statistical testing module helps to mitigate these challenges, resulting in a notable boost to the training process, as seen in the frugal setting. These findings suggest that our approach has the potential to improve instance segmentation methods for IRSTD, \addSylv{by combining object-level detection with fine-grained pixel-level representation}, which indicates a promising direction for future research.

\textit{Remark -- } The results presented in Table~\ref{tab:pixel_res} for SOTA segmentation methods may slightly differ from the results reported in the original papers. This variation stems from our model selection approach, which differs from the original studies. Specifically, we select the best-performing model based on validation set performance, whereas the original papers often optimize on the test set. By decoupling model selection from test set performance, we aim to ensure a fair and unbiased evaluation process, thereby avoiding overfitting to the test set.

\paragraph{Qualitative analysis}
Figure~\ref{fig:obj_scores_map} shows that our method produces remarkably clean objectness score maps compared to YOLOv7t and EFLNet, with only small target scores emerging from a near-zero background. The optimization process is key to this behavior, as it enables the network to adjust feature representations in a way that: 1)~targets are sufficiently distinct from the null hypothesis $H_0$, allowing for accurate detection, and 2)~background regions \addSylv{tend to} conform to $H_0$, mitigating the false alarms. This alignment, achieved through end-to-end training, is what makes our test operationally effective and robust: \addSylv{it allows for} high precision rates, even at low detection thresholds, as \addSylv{confirmed} by the precision curves provided in Figure~\ref{fig:pr_curves}. 
\addSylv{Crucially, this allows us to use a low, fixed threshold across all images, avoiding the need to manually tune it depending on image content or training conditions.}
This is a significant operational advantage over existing approaches\addSylv{, e.g. standard YOLO where the threshold often needs to be adjusted to prevent false alarms}, as it enables us to set a \addSylv{robust default value} with minimal risk of false alarms \addSylv{(since our method exhibits very few false alarms even at low thresholds)}.

\subsection{Greater robustness in challenging conditions}
\label{sec:robustness}

\begin{figure}[ht]
    \centering
        \begin{subfigure}[ht]{0.5\textwidth}
                \centering
                \includegraphics[width=0.9\textwidth]{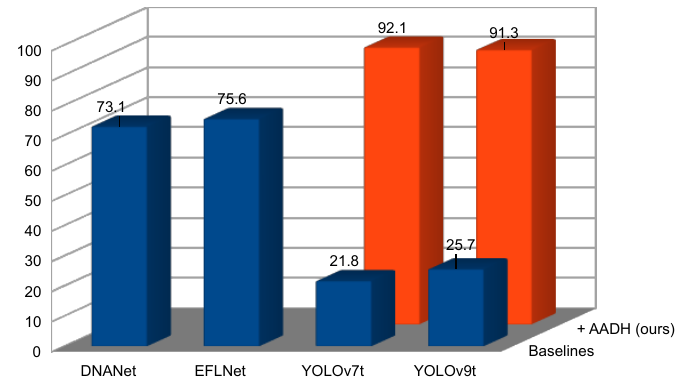}
                \caption{25-shot training}
                \label{fig:25-shot}
        \end{subfigure}
        
        \begin{subfigure}[ht]{0.5\textwidth}
                \centering
                \includegraphics[width=\textwidth]{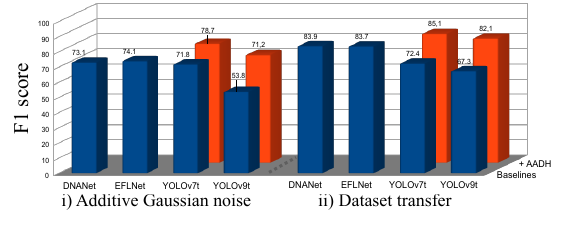}
                \caption{Noisy inference and dataset transfer robustness testing}
                \label{fig:gaussiannoise}
        \end{subfigure}
    
    \caption{Robustness towards few-shot training, inference on noisy data and transfer learning.}
    \label{fig:frugal_setting}
\end{figure}

\begin{figure*}
    \centering
    \includegraphics[width=0.87\linewidth]{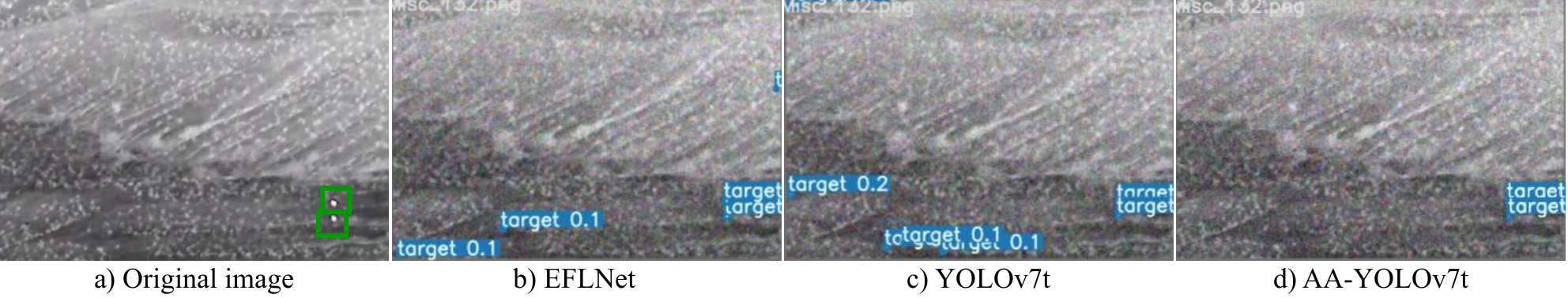}
    \caption{Detection results obtained on a particularly noisy image (with an additive Gaussian noise with spread $\sigma=0.1$).}
    \label{fig:ex_noise}
\end{figure*}

Evaluating the robustness of a method is important to ensure its reliability, accuracy, and safety in real-world applications, where data are often imperfect, scar\addSylv{c}e, and uncertain. To further evaluate the benefits of AA-YOLO, we assess its robustness under various challenging conditions.

\paragraph{Few-shot training} 
Figure~\ref{fig:25-shot} presents the results obtained when the compared methods are trained on only $10\%$ of the SIRST dataset (i.e., $25$ images). All our AA-YOLO variants (represented by dark orange bars) achieve \addSylv{strong performance}, close to those obtained with the full dataset. This robustness in frugal settings \addSylv{stems from} our explicit use of background information to discriminate target pixels, which helps \addSylv{compensate for} the lack of annotated data. In contrast, DNANet and EFLNet, although performing \addSylv{reasonably well}, do not match the performance of our AA-YOLO variants, while YOLO baselines struggle even more.

\paragraph{Inference on noisy data}
To evaluate robustness \addSylv{to noise}, we add Gaussian noise with \addSylv{standard deviation} $\sigma=0.1$ to our test set. As shown in~Figure~\ref{fig:gaussiannoise} i), integrating AADH into YOLO baselines significantly increases their robustness to nois\addSylv{y inputs}. Furthermore, our top-performing model, AA-YOLOv7t, outperforms EFLNet by over $4$ points in F1 score, demonstrating superior robustness to noisy inputs. Figure~\ref{fig:ex_noise} shows an example of inference on a particularly noisy data, with the absence of false alarms from our method highlighting the robustness of AADH in this context.

\paragraph{Transfer to another dataset}
\begin{figure}
    \centering
    \includegraphics[width=0.9\linewidth]{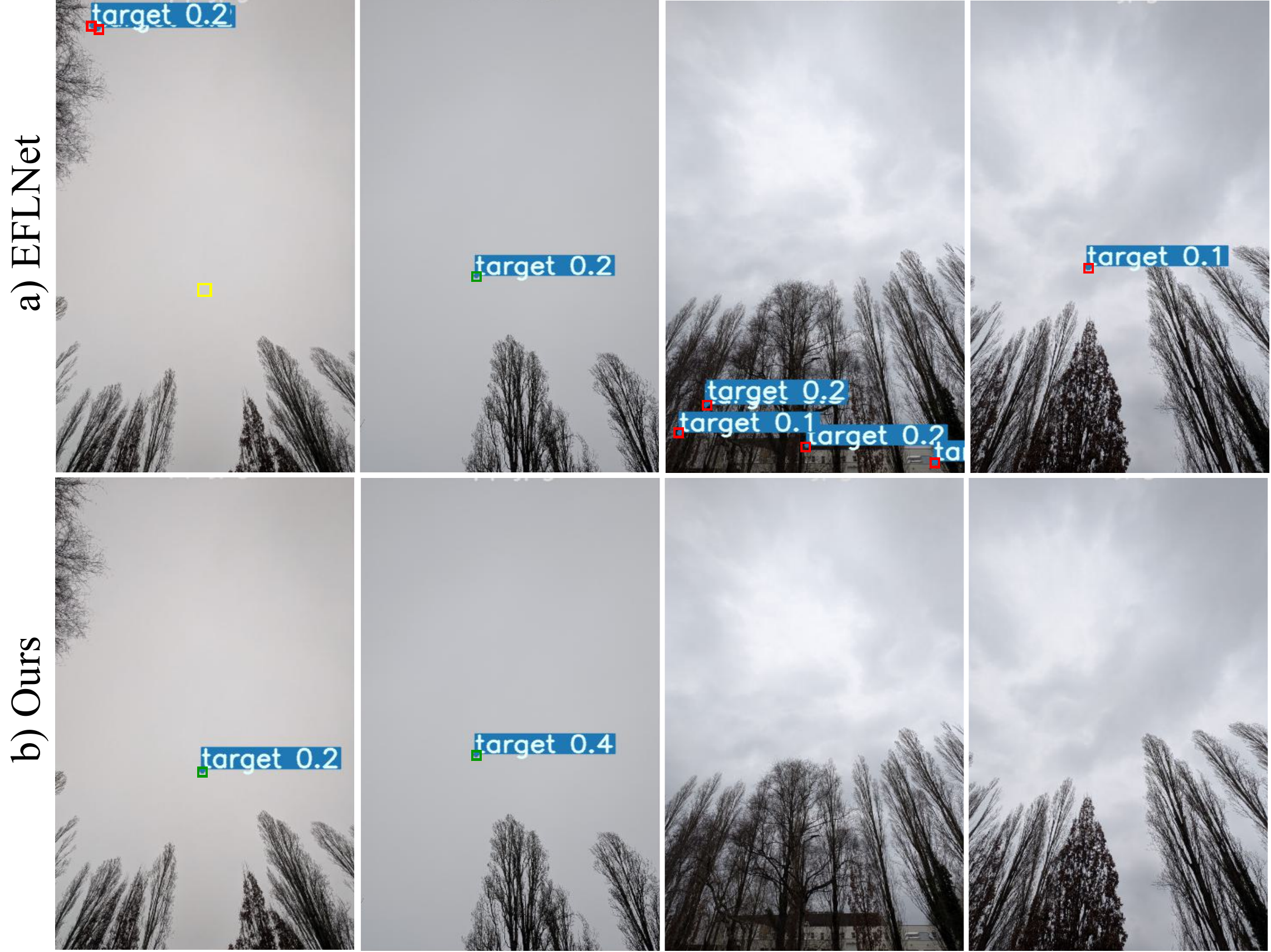}
    \caption{Results obtained in an operational context.}
    \label{fig:donnee_ops}
\end{figure}

We also assess the transferability of detectors from the SIRST dataset to the more challenging IRSTD-1k dataset. Figure~\ref{fig:gaussiannoise} ii) shows that our method enhances the performance of several YOLO baselines. For example, AADH improves YOLOv7t by $12.5$ points, achieving a F1 score that is only $7.4$ points \addSylv{below that obtained} when directly trained on the IRSTD-1k dataset.

\paragraph{Transfer to another modality}
Finally, we evaluate our algorithms in a real-world operational context using images captured with a mobile phone camera, which operates in the RGB modality and thus represents a different \addSylv{sensing} scenario. The full-resolution images are presented in~\ref{sec:high_res_data}; they \addSylv{depict} a sky scene, with a drone visible in the first two images and absent in the last two.
Figure~\ref{fig:donnee_ops} shows the detections produced by EFLNet and our best detector, AA-YOLOv7t, both set with a detection threshold of $0.1$. Our method successfully detects the drone in the first two images with \addSylv{no} false alarms. In contrast, EFLNet struggles to detect the drones and generates numerous false alarms. Although false alarms on trees may not be critical, the false alarm in the sky (last image) \addSylv{highlights} EFLNet's tendency to hallucinate targets. This shows the importance of assessing IRSTD algorithms on target-free \addSylv{scenes}.

\section{Ablation studies}
\label{sec:ablation}
This section provides the results of the ablation study performed on three components : 1) the formulation of the statistical test, 2) the number of channels in AADH, and 3) the parameter $\alpha$ in our activation function $\sigma_{\alpha}$. We also provide the computational consumptions of our methods.

\paragraph{Statistical testing formulation}
\label{sec:ablation_stat_test}
\bgroup
\def\arraystretch{1.15}
\begin{table}[] 
\small
  \centering
\begin{tabular}{lll} 
  \hline
     \multirow{2}{*}{\textbf{Statistical formulation}}  &  \multicolumn{2}{c}{\textbf{SIRST} }\\\cline{2-3}
      
        &  \textbf{F1 }  &$\textbf{AP}$ \\
     \hline
     $\mu_1 = \min \{X_1,...,X_C\}$ & 96.2 & 97.3 \\
    \cellcolor{green!5} $\mu_2 = \sum_{i=1}^C X_i$ & \cellcolor{green!5}\textbf{97.9} & \cellcolor{green!5}\textbf{98.5} \\

    Gaussian background hypothesis & \underline{97.6} & 98.3 \\
     
     \hline
     \multicolumn{3}{l}{\textbf{\# Channels in AADH}} \\
      
     1 & 92.3 &95.0 \\
     3 & 95.1 &97.3\\
     5 & 97.2 &98.3 \\
     6 & 97.2 & \underline{98.4} \\
     7 & \underline{97.6} & 98.3 \\
     
     \cellcolor{green!5}8 & \cellcolor{green!5}\textbf{97.9} & \cellcolor{green!5}\textbf{98.5} \\
     9 & 95.9 & 98.1 \\
     \hline
     \multicolumn{3}{l}{\textbf{Activation function - \textbf{$\alpha$}}} \\
     0.005 & 96.9 & 98.0\\
     \cellcolor{green!5}0.001 & \cellcolor{green!5}\textbf{97.9} & \cellcolor{green!5}\textbf{98.5} \\
     0.0005 & 97.1 &\underline{98.4} \\
    
     \hline
  \end{tabular}
  \caption{Ablation on the statistical testing formulation, using YOLOv7t backbone.}
  \label{tab:ablation_stat}
\end{table}
\egroup

Table~\ref{tab:ablation_stat} compares three statistical formulations for AADH using the YOLOv7t backbone: the formulation using the $\mu_1$ measure, the one using the $\mu_2$ measure, and the Gaussian background hypothesis.

The comparison clearly shows that $\mu_2$ significantly outperforms $\mu_1$: the F1 score improves by 1.7\%. This superior performance can be attributed to the limitations of $\mu_1$, which assumes that all channels contain information about the target once one channel does. This assumption hinders the extraction of well-diversified features by the network, thereby reducing the quality of the extracted features. We also considered an alternative hypothesis, namely that the background follows a Gaussian distribution. Knowing that the sum of $C$ independent standard normal random variables follows a $\chi^2$ distribution with $C$ degrees of freedom, the testing function evaluated on $X_k$ boils down to $ \frac{\Gamma(\frac{C}{2},\frac{1}{2} ||X_k||^2_2)}{\Gamma(C/2)} $. However, this assumption leads to slightly inferior detection performance, suggesting that the exponential assumption better aligns with the structure of the latent features.

\paragraph{Number of channels in AADH}
\label{sec:ablation_numchans}

Figure~\ref{fig:F_variations} illustrates \addSylv{the} variations of $F$ and $-\ln{F}$ \addSylv{for} different values of $C$, where the choice of $C$ \addSylv{influences} the sharpness of $F$. We therefore perform an ablation study on the parameter $C$ in the AADH using YOLOv7t backbone. The results presented in Table~\ref{tab:ablation_stat} show that very low values of \( C \) result in poor performance, while excessively high values (above 8) lead to training difficulties and make it harder for the network to converge. To balance performance and stability, we recommend setting the number of channels \( C \) between 5 and 8, as this range avoids the aforementioned issues and ensures optimal results.

\paragraph{$\alpha$ parameter in $\sigma_{\alpha}$ activation function}
The last three lines of Table~\ref{tab:ablation_stat} show that the choice of $\alpha$ has an impact on the performance. Specifically, setting $\alpha=0.001$ leads to optimal performance when using AA-YOLOv7t.

\bgroup
\def\arraystretch{1.15}
\begin{table}[h] 
\small
  \centering
\begin{tabular}{llcllllllllllll} 
  \hline
    \textbf{Method} & \textbf{\#P (M)$\downarrow$} & \textbf{GFLOPs$\downarrow$} \\
     \hline
    DNANet & 4.7 & 14.26 \\
    SCTransNet & 11.2 & 20.24\\
    EFLNet & 38.3 & 102.2 \\
     \hline
    YOLOv7+PConv+SD & 6.0 & 14.0\\
    AA-YOLOv7+PConv+SD & 6.2 & 14.7\\
     \hline
    YOLOv5-seg & 7.4 & 25.7\\
    AA-YOLOv5-seg & 7.6 & 26.3\\
     \hline
    YOLOv7 & 37.3 & 103.2 \\
    AA-YOLOv7 & 37.5 & 104.5 \\
     \hline
    YOLOv7t & 6.0 & 13.0\\
    AA-YOLOv7t & 6.2 & 13.7\\
     \hline
    YOLOv9t & 1.4 & 5.0\\
   
    AA-YOLOv9t & 1.5 & 5.3\\
     \hline
    
     \hline
             
  \end{tabular}
  \caption{Number of training parameters (\#P, expressed in million (M)) and GFLOPs for the evaluated methods.}
  \label{tab:comput_res}
\end{table}
\egroup

\paragraph{Computational consumption}
Table~\ref{tab:comput_res} presents a computational analysis of our methods, including the number of model parameters and floating-point operations (FLOPs) expressed in GFLOPs. Our results show that adding the AADH module to the YOLO baseline incurs a negligible increase in parameters (approximately 0.2M) and FLOPs (around 5\%). This suggests that our approach maintains the deployability of the baseline YOLO model while significantly improving performance and robustness under challenging conditions. Notably, our AA-YOLOv7t outperforms EFLNet, which has 6 times more parameters and 7 times more GFLOPs. Furthermore, when combined with a lighter backbone (such as YOLOv9t), our approach achieves comparable performance to EFLNet while using 25 times fewer parameters and 19 times fewer GFLOPs. This substantial reduction in computational cost makes our method an attractive solution for real-world applications.

\subsection{Discussion about the versatility of our method}

We have proposed a novel detection head, AADH, specifically designed for IRSTD. As demonstrated in the previous paragraphs, our method not only makes any YOLO\addSylv{-based} model competitive with SOTA detectors\addSylv{,} but also significantly enhances robustness under challenging conditions, \addSylv{thereby} increasing \addSylv{its} trustworthiness. The frugal nature of our contribution -- a \addSylv{lightweight} module added \addSylv{at} the end of a YOLO network -- \addSylv{enables seamless integration into} various \addSylv{architectures} and constraints.

\paragraph{Task versatility}
\addSylv{A particularly noteworthy aspect of our approach is its versatility across tasks. Since our method relies} solely on shape and pattern analysis, independent of the physical priors specific to IR imaging\addSylv{, it appears naturally suited} to other tiny object detection tasks beyond IRSTD.
For example, in remote sensing, detecting small objects such as vehicles in complex environments is a common challenge. This task differs from IRSTD \addSylv{by including a larger number of} objects, some of \addSylv{which are} larger and better resolved. \addSylv{It is therefore legitimate to} question whether IRSTD methods can \addSylv{generalize} effectively to such tasks. 
To address this question, we evaluated the effectiveness of our approach on the VEhicule Detection in Aerial Imagery dataset (VEDAI,~\cite{razakarivony2016vehicle}), which consists of $1200$ IR satellite images with a resolution of $512\times 512$ pixels. 
The results\addSylv{, presented} in Table~\ref{tab:VEDAI}\addSylv{,} show that YOLO baselines perform poorly on this task, despite their simplicity and \addSylv{operational appeal}. \addSylv{In contrast, integrating} our module \addSylv{in}to any YOLO detector, regardless of its size or complexity, yields highly competitive performance compared to EFLNet with significantly fewer training parameters, while demonstrating robustness in various scenarios such as frugal setting. 
\addSylv{These findings suggest that} our method can be confidently employed to achieve excellent performance in various small object detection tasks, particularly \addSylv{in settings with limited data or resources}.

\paragraph{Architectural versatility}

Our modular design provides \addSylv{architectural} flexibility \addSylv{enabling} us to \addSylv{address a wide range of} challenges. By combining lightweight networks (e.g., YOLOv9t) with our AADH module, we can achieve high performance even in resource-constrained settings. 
Moreover, \addSylv{the modularity of our approach facilitates adaptation} to different tasks, such as instance segmentation, which helps overcome \addSylv{common issues such as target} adjacency and fragmentation. This \addSylv{results in accurate} pixel-level predictions while maintaining object-level accuracy.
\addSylv{Our design is also well-suited to scenarios involving architectural constraints.} For example, when training conditions are very difficult, it is \addSylv{often beneficial} to rely on pre-trained weights for the encoder to enhance the performance and robustness of the detectors. However, off-the-shelf pre-trained weights are only available for \addSylv{standard} encoders such as ResNet50, ViT-B, or Swin Transformers.
Unlike \addSylv{many} IRSTD\addSylv{-specific} methods in the literature that \addSylv{rely on custom} encoders, we can easily add AADH to a YOLO with a generic encoder such as ResNet50 or Swin, and thus consider pre-trained weights. This advantage is particularly significant \addSylv{in light of recent advances in} self-supervised learning\addSylv{, which have demonstrated considerable impact across vision tasks}.
 
\paragraph{Limits and perspectives}
\label{sec:limits_vedai}
\begin{figure}
    \centering
    \includegraphics[width=1\linewidth]{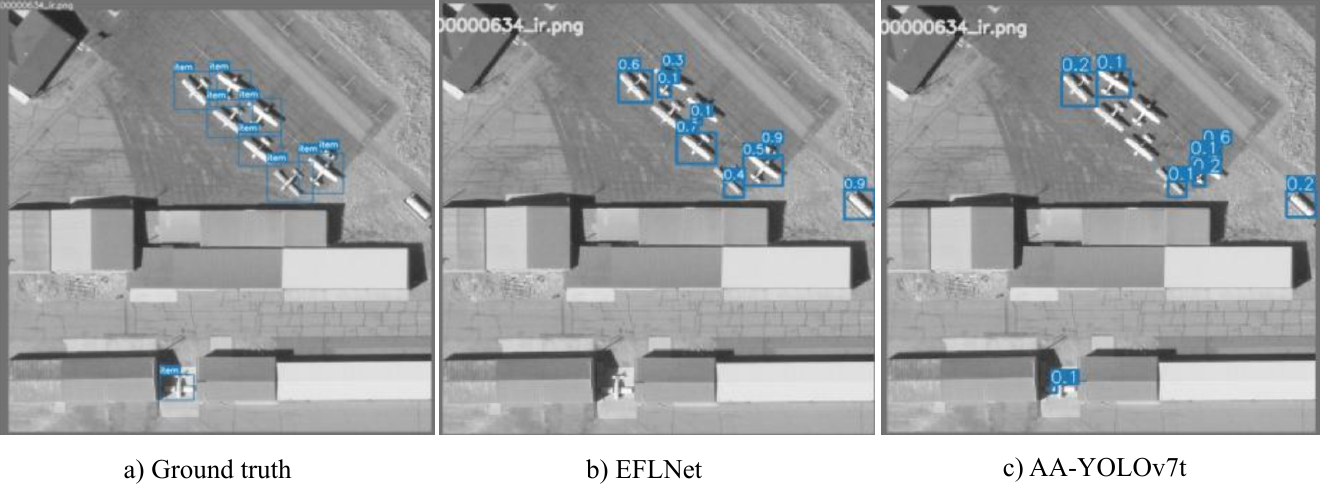}
    \caption{Illustration of the limits of our method on large, numerous vehicle detection. The image is extracted from the VEDAI dataset.}
    \label{fig:vedai_large}
\end{figure}

\addSylv{While o}ur method is \addSylv{well-suited for} the detection of \textit{unexpected} and rare events, it has limitations when it comes to detecting large and numerous objects. As \addSylv{shown in} Figure~\ref{fig:vedai_large}, \addSylv{AA-YOLO tends to} underdetect large and multiple planes in the VEDAI dataset compared to SOTA methods. This \addSylv{outcome is consistent} with theoretical expectations\addSylv{:} such objects no longer \addSylv{qualify as statistical anomalies with respect to the background and thus fall outside the scope of our anomaly-based detection strategy}.
\addSylv{While the statistical test relies on simplifying assumptions, it achieves robust performance in practice and avoids overfitting to spurious patterns, as shown by our ablations and transfer scenarios. More sophisticated probabilistic modeling is left as future work.}

\bgroup
\def\arraystretch{1.15}
\begin{table}[t] 
\small
  \centering
\begin{tabular}{lccc} 
  \hline
     \multirow{2}{*}{\textbf{Method}} & \multirow{2}{*}{\textbf{\#P (M)}} &  \multicolumn{2}{c}{\textbf{VEDAI} } \\\cline{3-4}
      
      &  &  \textbf{F1 }  &$\textbf{AP}$  \\
     \hline
     
    EFLNet & 38.3 & \underline{82.8} & \textbf{87.2} \\
    
    YOLOv7 & 37.3 & 80.5 & 84.8 \\
    YOLOv7t & 6.0 & 80.9 &85.3 \\
    YOLOv9t & 1.4 & 76.9 & 81.7 \\
    AA-YOLOv7 & 37.5 & 82.0& 86.1 \\
    AA-YOLOv7t & 6.2 & \textbf{83.0}& \underline{87.0 }\\
    AA-YOLOv9t & 1.5 & 80.5 & 85.7 \\
     \hline   
             
  \end{tabular}
  \caption{Results obtained on the VEDAI dataset. \#P is the number of training parameters, in million (M).}
  \label{tab:VEDAI}
\end{table}
\egroup

\section{Conclusion}

In this paper, we have presented \addSylv{AA-YOLO,} a simple \addSylv{yet effective} method to improve the detection of small targets in IR images. Our approach involves integrating a statistical anomaly testing \addSylv{directly} into the detection head of YOLO-type networks, thereby \addSylv{enabling the detection of} targets as \addSylv{deviation from} the background \addSylv{distribution}.
Our method not only improves any YOLO detector for small object detection but also leads to SOTA results on two \addSylv{widely used} IRSTD benchmarks. Last but not least, our AA-YOLO demonstrates a robustness that is all the more impressive in few-shot training (achieving $90\%$ of the \addSylv{full-}performance SIRST dataset), and the versatility of our approach allows it to adapt to various scenarios with severe resource constraints.

\textbf{Acknowledgments --} This project was provided with computing HPC and storage resources by GENCI at IDRIS thanks to the grant 2024-AD011015959 on the supercomputer Jean Zay's V100 and A100 partitions.

\bibliographystyle{elsarticle-num} 

\bibliography{article}

\newpage
\newpage

\appendix
\setcounter{figure}{0}
\setcounter{table}{0}
\renewcommand{\thefigure}{S\arabic{figure}}
\renewcommand{\thetable}{S\arabic{table}}

{ \centering \LARGE \textbf{Anomaly-Aware YOLO: A Frugal yet Robust Approach to Infrared Small Target Detection}\\[1em]
  \Large \textbf{Supplementary Materials}\\[5em]}
    This document is the supplementary material of the
paper entitled "Anomaly-Aware YOLO: A Frugal yet Robust Approach to Infrared Small Target Detection". It provides additional results and visualizations, as well as the details on the training parameters used for all methods.

\section{Additional results and metrics}
\label{sec:additional_res}

\bgroup
\def\arraystretch{1.15}
\begin{table*}[h] 
\scriptsize
  \centering
\begin{tabular}{p{1.6cm}lcllllllllllll} 
  \hline
     \multirow{2}{*}{\textbf{Backbone}} &  \multirow{2}{*}{\textbf{AADH}}  &  \multicolumn{6}{c}{\cellcolor{blue!5}\textbf{SIRST} } & \multicolumn{6}{c}{\cellcolor{red!5}\textbf{IRSTD-1k} }\\\cline{3-14}
      
     &   &  \textbf{F1 $\uparrow$}  &$\textbf{AP$\uparrow$}$& \textbf{P $\uparrow$}  &$\textbf{R $\uparrow$}$&\textbf{Fa $\downarrow$ }  &\textbf{AP$_{s}$ $\uparrow$}  & \textbf{F1 $\uparrow$}  & $\textbf{AP$\uparrow$}$ & \textbf{P $\uparrow$}  &$\textbf{R$\uparrow$}$&\textbf{Fa $\downarrow$}  &\textbf{AP$_{s}$ $\uparrow$}     \\
     \hline
     ACM& & 95.4 & 95.2 & 95.6 & 95.2 & 0.07 & 87.1 & 90.9 & 88.2 & 92.7 & 89.2 & 0.11 & 77.8\\
     AGPCNet & & 93.8 & 92.2 & 93.1 & 94.5 & 0.08 & 93.0 & 91.1 & 88.9 & 92.1 & 90.2 & 0.11 & 74.1 \\
     DNANet & & 97.1&98.4 &96.9 &97.3 &0.04 &96.1 & 90.7 & 87.0 & 90.6 & 90.9 & 0.14 & 79.1 \\ 
    
     RDIAN & & 95.9 & 93.8 & 93.9 & \textbf{98.2} & 0.08 & 86.3 & 86.7 & 84.5 & 88.7 & 84.8 & 0.16 & 67.3 \\
     SCTransNet & & 95.4 & 95.9 & 93.1 & 97.9 & 0.08 & 92.9 & 91.9 & 90.8 & 92.1 & \underline{91.7} & 0.13 & 87.0 \\
     SIRST-5K & & 96.8 & 98.1 & 97.9 & 95.9 & 0.02 &95.2 & 90.1 &89.3 & 90.8 & 89.7 & 0.14 & 86.7\\
     MSHNet & & 94.8 & 95.6 & 93.9 & 95.9 & 0.07 & 92.3 & 92.0 & 91.1 & 93.0 & 91.1 & 0.10 & 88.8 \\
      DATransNet& & 93.5& 92.3& 94.4&92.7 &0.07 &74.9 & 89.1 & 86.7 & 87.1 & 91.2 & 0.20 &  73.2\\
     EFLNet  & & 96.9 & 98.3 & 97.9 & 95.9 & 0.02 & 97.6 & \underline{92.5} & \underline{96.5} & 93.6 & 91.4 & 0.09 & \underline{89.8} \\
     YOLOv7t +PConv+SD & & 96.8 &97.8 & \underline{98.9} & 94.8 & \underline{0.01} & 97.1 & 87.8 & 93.2 & 92.7 & 83.4 & 0.10 & 86.1 \\
     \hline
     YOLOv5-seg & & 95.8 & 98.2 & 96.8 & 94.8 & 0.04 & 97.6 & 87.6 & 91.9 & 90.1 & 85.1 & 0.14 & 83.7 \\
     YOLOv5-seg  & \checkmark & 96.9 & \textbf{98.6} & 96.9 & 96.9 & 0.04 & 97.8 & 90.2 & 88.9 & 89.3 & 91.1 & 0.16 & 82.1\\
     \hline
     YOLOv7  & & 96.9 & 98.2 & 96.0 & \underline{97.9} & 0.05 & 97.1 &  \textbf{92.6} & 95.2 & 93.0 & \textbf{92.3} & 0.10 & 87.1 \\
     YOLOv7 & \checkmark & \textbf{97.9} & 98.5 & 97.9 & \underline{97.9} & 0.02 & \textbf{98.0} & 91.2 & 95.6 & 93.1 & 89.4 & 0.09 & 86.3\\
     \hline
     YOLOv7t +PConv+SD & & 96.8 &97.8 & \underline{98.9} & 94.8 & \underline{0.01} & 97.1 & 87.8 & 93.2 & 92.7 & 83.4 & 0.10 & 86.1\\
     YOLOv7t +PConv+SD & \checkmark &  96.9 & 98.0 & 96.9 & 96.9 & 0.04 & 97.0 & 91.1 & 96.0 & 91.1 & 91.1 & 0.13 & 89.5\\
     \hline
     YOLOv9t & & 94.5 & 97.0 & 92.2 & 96.9 & 0.09 & 97.3 & 90.9 & 95.4 & \underline{94.0} & 88.1 & \underline{0.08} & 83.9 \\
     YOLOv9t & \checkmark &\underline{97.4} & \textbf{98.6} & \underline{98.9} & 95.9 & \underline{0.01} & 97.8 & 91.4 & 95.8 & 93.4 & 89.4 & 0.09 & 87.9\\
     \hline
     YOLOv7t & & 96.3& 98.2 & \underline{98.9} & 93.8 & \underline{0.01} & 97.1 & 85.3 & 91.5 & 84.4 & 86.1 & 0.24 & 83.6\\
     YOLOv7t &\checkmark & \textbf{97.9} & \underline{98.5} & \textbf{100.0} & 95.9 & \textbf{0.00} & \underline{97.9} & \underline{92.5} & \textbf{96.6} & \textbf{95.1} & 90.1 & \textbf{0.07} & \textbf{90.9} \\
     \hline
    
     \hline
             
  \end{tabular}
  \caption{Additional results and metrics obtained on SIRST and IRSTD-1k datasets. The best results are in bold, and the second best results are underlined.}
  \label{tab:ablation_AADH}
\end{table*}
\egroup

In order to gain a deeper understanding of the methods' sensitivity and detection capability, we provide their precision, recall, and false alarm rates per image on both SIRST and IRSTD-1k datasets in Table~\ref{tab:ablation_AADH}. Our findings show that our best method AA-YOLOv7t leads to the lowest false alarm rate. More specifically, on the SIRST dataset, AADH enhances precision, while on the IRSTD-1k dataset, it leads to a substantial increase in recall, which is a challenging metric for baseline methods. Importantly, this increase in recall comes at a very small cost to precision, showing that AADH demonstrates a good balance between precision and recall.

\section{High-resolution operational data}
\label{sec:high_res_data}
\begin{figure*}
    \centering
    \includegraphics[width=\linewidth]{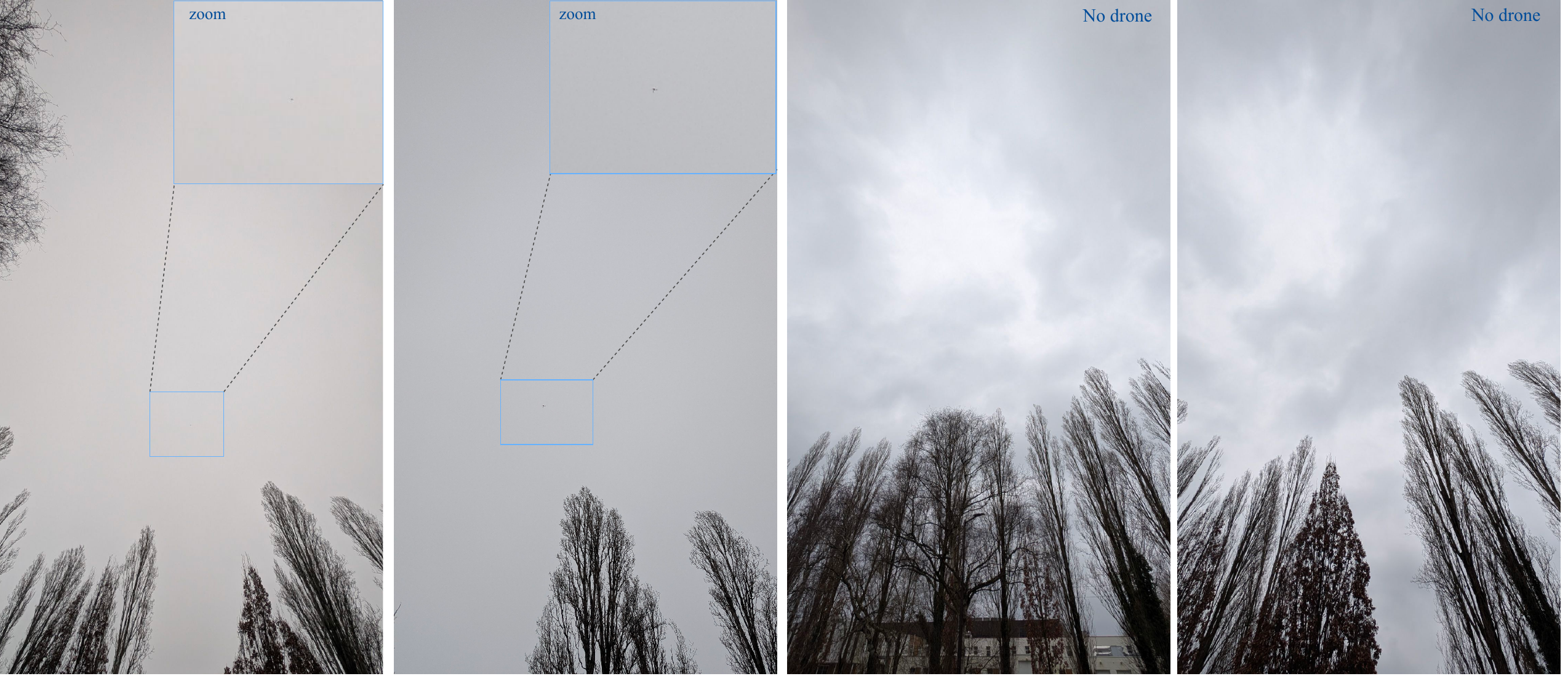}
    \caption{High-resolution operational data. The first two images contain a drone that can be better seen in the zoomed-in view in the top-right corner, whereas the last two images are target-free.}
    \label{fig:im_op_origin}
\end{figure*}

Figure~\ref{fig:im_op_origin} displays the data used to evaluate the robustness of our algorithm under operational conditions, shown here in high resolution. Notably, the drone in the first image is nearly imperceptible to the naked eye, making it a challenging target to detect.

\section{Training parameters}
\label{sec:training_param}
\bgroup
\def\arraystretch{1.15}
\begin{table*}[h] 
\scriptsize
  \centering
\begin{tabular}{p{4.5cm}ll} 
  \hline
  
  \hline
     Architecture & Parameter & Value \\
      \hline
    \multirow{5}{=}{(AA-)YOLOv7, (AA-)YOLOv7t, (AA-)YOLOv5-seg, \\ (AA-)YOLOv7t+PConv+SD} & LR & 0.01 \\\cline{2-3}
    & optimizer & SGD  - weight decay:  0.0005 \\\cline{2-3}
     & scheduler  & OneCycleLR - final LR: 0.1 \\\cline{2-3}
    & epochs & 600 \\\cline{2-3}
     & batch size & 16 \\\cline{2-3} 
   \hline
   
     \hline
        \multirow{5}{=}{(AA-)YOLOv9, (AA-)YOLOv9t} & LR & 0.01 \\\cline{2-3}
         & optimizer & SGD  - weight decay: 0.0005 \\\cline{2-3}
     & scheduler  & OneCycleLR - final LR: 0.01 \\\cline{2-3}
    & epochs & 600 \\\cline{2-3}
   & batch size & 16 \\
     \hline

     \hline
        \multirow{5}{=}{EFLNet} & LR & 0.01 \\\cline{2-3}
        & optimizer & SGD - weight decay: 0.0005 \\\cline{2-3}
   & scheduler  & OneCycleLR - final LR: 0.1 \\\cline{2-3}
    & epochs & 600 \\\cline{2-3}
     & batch size & 16 \\\cline{2-3} 
     
    & other & focal loss - gamma: 1.0 \\
     \hline

\hline
     
        \multirow{5}{=}{ACM, AGPCNet, DNANet, RDIAN, DATransNet} & LR & 0.0005 \\\cline{2-3}
         & optimizer & Adam \\\cline{2-3}
     & scheduler & MultiStepLR (steps: [200, 300], gamma: 0.1) \\\cline{2-3}
    
    & epochs & 400 \\\cline{2-3}
    
    & batch size & 4 \\\cline{2-3} 
     \hline

\hline
     
        \multirow{5}{=}{SCTransNet} & LR & 0.001 \\\cline{2-3}
         & optimizer & Adam \\\cline{2-3}
     & scheduler & CosineAnnealingLR - $\eta_{min}$: $10^{-5}$ \\\cline{2-3}
    
    & epochs & 1000 \\\cline{2-3}
    
    & batch size & 4 \\\cline{2-3} 
    \hline

    \hline
     
        \multirow{5}{=}{SIRST-5K} & LR & 0.05 \\\cline{2-3}
         & optimizer & Adagrad \\\cline{2-3}
     & scheduler & CosineAnnealingLR  - $\eta_{min}$: $10^{-5}$ \\\cline{2-3}
    
    & epochs & 1500 \\\cline{2-3}
    
    & batch size & 4 \\\cline{2-3} 
    \hline

    \hline
     
        \multirow{5}{=}{MSHNet} & LR & 0.05 \\\cline{2-3}
         & optimizer & Adagrad \\\cline{2-3}
     & scheduler & \textit{NA} \\\cline{2-3}
    
    & epochs & 400 \\\cline{2-3}
    
    & batch size & 4 \\\cline{2-3} 
    \hline
  \end{tabular}
  \caption{Training configurations used for SIRST, IRSTD-1k and VEDAI datasets.}
  \label{tab:training_params}
\end{table*}
\egroup

Table~\ref{tab:training_params} provides the training configurations used for training SOTA segmentation networks, YOLO detectors and our methods on SIRST, IRSTD-1k and VEDAI datasets. The following sources were used for model training: 
\begin{itemize}
    \item EFLNet, SCTransNet, DATransNet, SIRST-5K and MSHNet were trained using the original repository provided by the authors.
    \item For ACM, AGPCNet, and DNANet, we utilized the implementations and training parameters provided by DATransNet GitHub repository.
    \item YOLOv7 and YOLOv9-based methods were trained using the following repository: https://github.com/WongKinYiu/yolov7.
    \item (AA-)YOLOv5-seg networks were trained using the following repository: https://github.com/ultralytics/yolov5.
\end{itemize}
\end{document}